%% file: main.tex
\documentclass[a4paper, conference, 10pt, twocolumn]{IEEEtran}

\usepackage{times}
\usepackage{balance}

\usepackage{amsmath,graphicx,rotating,xcolor}
\usepackage{hyperref}
\usepackage[symbol]{footmisc}

\def \etal {\emph{et al.}}

\newcommand\crule[3][black]{\textcolor{#1}{\rule{#2}{#3}}}

\definecolor{rgb_color_unlabeled}{RGB}{0,0,0}
\definecolor{rgb_color_car}{RGB}{100,150,245}
\definecolor{rgb_color_motorcyclist}{RGB}{150,30,90}
\definecolor{rgb_color_road}{RGB}{255,0,255}
\definecolor{rgb_color_parking}{RGB}{255,150,255}
\definecolor{rgb_color_sidewalk}{RGB}{75,0,75}
\definecolor{rgb_color_building}{RGB}{255,200,0}
\definecolor{rgb_color_fence}{RGB}{255,120,50}
\definecolor{rgb_color_vegetation}{RGB}{0,175,0}
\definecolor{rgb_color_trunk}{RGB}{135,60,0}
\definecolor{rgb_color_terrain}{RGB}{150,240,80}
\definecolor{rgb_color_pole}{RGB}{255,240,150}
\definecolor{rgb_color_traffic-sign}{RGB}{255,0,0}

\begin{document}

\title{HPGNN: Using Hierarchical Graph Neural Networks for Outdoor Point Cloud Processing}

\author{Arulmolivarman Thieshanthan, Amashi Niwarthana, Pamuditha Somarathne,\\ Tharindu Wickremasinghe, Ranga Rodrigo\\
\emph{Department of Electronics and Telecommunication Engineering, University of Moratuwa, Sri Lanka}\\

\author{\IEEEauthorblockN{Arulmolivarman Thieshanthan\IEEEauthorrefmark{1}\IEEEauthorrefmark{2},
Amashi Niwarthana\IEEEauthorrefmark{1}\IEEEauthorrefmark{2},
Pamuditha Somarathne\IEEEauthorrefmark{1}\IEEEauthorrefmark{2},\\
Tharindu Wickremasinghe\IEEEauthorrefmark{1}\IEEEauthorrefmark{2},
Ranga Rodrigo\IEEEauthorrefmark{1}}

\IEEEauthorblockA{\IEEEauthorrefmark{1}Department of Electronic and Telecommunication Engineering, University of Moratuwa, Sri Lanka}}

}

\maketitle
\thispagestyle{empty}

\begin{abstract}
Inspired by recent improvements in point cloud processing for autonomous navigation, we focus on using hierarchical graph neural networks for processing and feature learning over large-scale outdoor LiDAR point clouds. We observe that existing GNN based methods fail to overcome challenges of scale and irregularity of points in outdoor datasets. Addressing the need to preserve structural details while learning over a larger volume efficiently, we propose Hierarchical Point Graph Neural Network (HPGNN). It learns node features at various levels of graph coarseness to extract information. This enables to learn over a large point cloud while retaining fine details that existing point-level graph networks struggle to achieve. Connections between multiple levels enable a point to learn features in multiple scales, in a few iterations.
We design HPGNN as a purely GNN-based approach, so that it offers modular expandability as seen with other point-based and Graph network baselines. To illustrate the improved processing capability, we compare previous point based and GNN models for semantic segmentation with our HPGNN, achieving a significant improvement for GNNs (+36.7 mIoU) on the SemanticKITTI dataset.
\footnote{\IEEEauthorrefmark{2} These authors contributed equally to this work.}

\end{abstract}

\input{HPGNN/1_Introduction}
\input{HPGNN/2_HPGraph}
\input{HPGNN/3_Experiments_n_Discussions}
\input{HPGNN/4_Conclusion}
\input{HPGNN/Acknowledgement}





\pagebreak
\newpage

\bibliographystyle{IEEEtran}
\bibliography{main}


\end{document}

%% file: HPGNN/1_Introduction.tex
\section{Introduction}\label{section1}

We focus on feature learning in outdoor LiDAR point clouds, an important perception manner commonplace in vision based autonomous navigation. A setup for collecting sequential scans to form point clouds includes a 3D LiDAR scanner mounted on a vehicle  \cite{Hungryfor3DLIDAR2020}. One challenge of such large-scale outdoor LiDAR point clouds is the high volume of points; generally millions of points per frame of observation \cite{behleySemanticKITTI,geigerSemantic3D}. Computing features for each point, with redundancies at dense regions of the cloud, poses a significant computational burden. Another challenge is the irregular distribution of points. Due to the spreading of the detecting beam from the LiDAR source, the observed cloud is denser in the immediate neighbourhood, and sparser with radial distance from the source. In processing such unordered points, repeated grouping and sampling of large clouds add to the complexity.\
To alleviate such challenges, a suitably down-sampled graphical representation to learn point features through a Graph Neural Network (GNN) is an idea explored \cite{ShiPointGNN}. At each node, these features are refined to improve the semantic feature representation. During a GNN iteration, the receptive field of the point grows, and its feature is updated from the aggregation of point features in the receptive field \cite{GRLBook}. The problem with this approach is its poor scalability for large outdoor point clouds. We incorporate the idea of hierarchical learning and feature pyramid schemes \cite{GraphFPN} to avoid the need of many iterations to expand the receptive field. 
\input{images/image1}
\label{sec:related}

To make feature learning over a large graph computationally efficient, voxelization \cite{ShiPointGNN, ZhuCylindricalAsym}  is a common approach. Such methods vary the sampling scheme based on the structure of the observed point cloud. As shown in Figure \ref{fig:prev} (a) they form a coarse level graph representation, learn features from the GNNs, and interpolate them onto the original cloud.

Graph coarsening strategies for reducing computation must strike a balance by sacrificing the accuracy of segmentation \cite{HuangGraphCoarsening}. Achieving an optimum coarseness for downsampling is a hard task, since it depends heavily on the dataset and the scale of the objects that we wish to segment. Due to these reasons, Graph Network based attempts that have been made to process large scale outdoor point clouds at the point-scale have thus far not been on par with other approaches.

To address these challenges, we propose a GNN representation---Hierarchical Point GNN (HPGNN)---that can be extended with modularity for feature encoding, and point-cloud processing tasks. We use scalable voxelizations to form graph connections at varying levels of coarseness (Figure \ref{fig:vox}). We extend the ideas of downsampling and graph coarsening \cite{ZhuCylindricalAsym} to address the scaling of graph structured data for large outdoor point clouds.

HPGNN preserves finer details, while allowing to learn across a larger scale. It leverages existing proximity relations between the points and their features to learn both contextual information locally, as well as spatial structural information in a global scope. As shown in Figure \ref{fig:prev} (b), given sufficient learning iterations (T1, T2) at each scale, connections between two hierarchical levels enable a more rich feature representation.

Our main contribution is incorporating a hierarchical learning scheme to improve semantic segmentation of large scale outdoor point-clouds, with computational feasibility in few iterations. We conduct experiments on SemanticKITTI \cite{behleySemanticKITTI} and nuScenes-Lidarseg \cite{fong2021panopticnuscenes} outdoor datasets  to evaluate the proposed architecture. HPGNN outperforms other GNN-based frameworks and other point-based schemes in single scan semantic segmentation.

%% file: images/image1.tex
\begin{figure}[t]
\centering
   \includegraphics[width=0.85\linewidth]{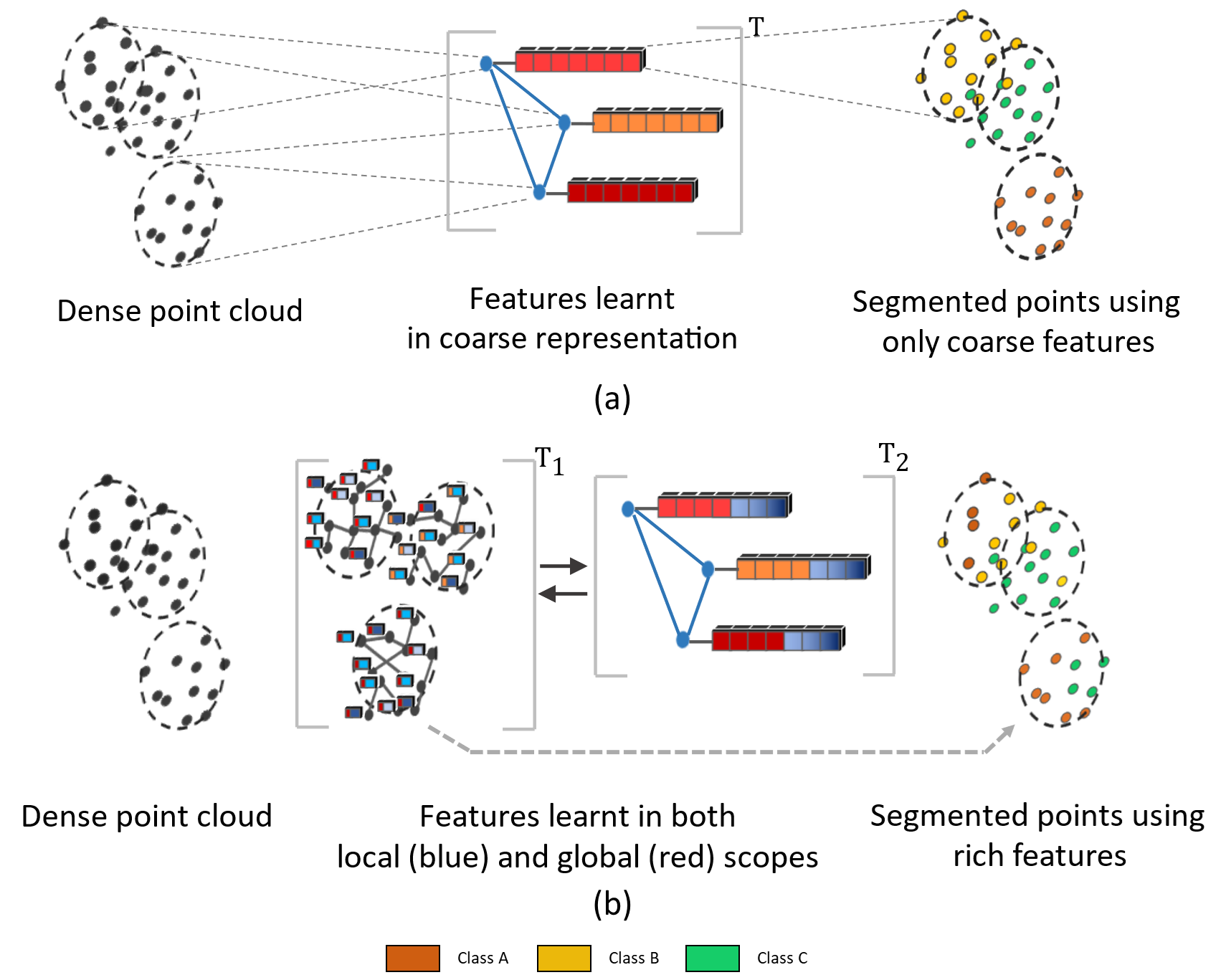}
   \caption{Acquiring rich features using two levels of coarseness. a) A coarsening approach learns features at one scale, using downsampled points. Features loose fine information. b) A mixture of coarse and fine features and passing between the levels to create rich features. HPGNN follows this approach to segment with higher resolution.}
\label{fig:prev}
\end{figure}

%% file: HPGNN/2_HPGraph.tex
\input{images/image2}
\input{images/image3}

\section{HPGNN for Semantic Segmentation}\label{section3}

\subsection{Voxelization and Keypoint Selection}\label{section3.1}
A common approach to reduce the point cloud density before processing is to use a voxel grid. PointGNN \cite{ShiPointGNN}, SSCN \cite{SSCN}, SegCloud \cite{TchapmiSEGCloud} and GridGCN \cite{XuGridGCN}, use cubic voxel grids. They do not exploit the irregular distribution within a point cloud. Zhu \etal \cite{ZhuCylindricalAsym} propose a cylindrical voxel partition with asymmetric 3D convolutions as an improvement on regular voxel-based models. These methods voxelize the clouds at a single scale. As the point cloud gets larger and irregular, voxelization schemes to prevent abstracting out fine details are not explored.

 We exploit the inherent cylindrical nature of LiDAR data following \cite{ZhuCylindricalAsym}, to parametrise the 3D space using cylindrical coordinates ($r, \theta, z$). We create a scalable voxelization scheme where linear intervals of adjustable magnitude along the radial distance, azimuth angle, and vertical height define voxel boundaries. This creates larger voxels as the radial distance increases. In a sense, the voxel size defines the “sieve” through which we filter the highly dense LiDAR points. The voxelization is parametrised such that each cylindrical block corresponds to a representative ``key point". A key point is a point with the mean position of the points inside a given voxel. The key points thus formed will create the nodes of the subsequent graph.\ 

Consider two cylindrical grids parametrised by ($r_L, \theta_L, z_L$) and ($r_H, \theta_H, z_H$); each with a different coarseness to sample the point cloud (Figure \ref{fig:vox} (b), (c)). They form nodes of two graphs, corresponding to two levels of coarseness (Figure \ref{fig:vox} (d)). We maintain connections between the two graphs so that each point in a finer layer may learn from a wider scope through its connection to the coarser layer.
This enables to process a wider scope of the point cloud, while preserving structural information that would otherwise be lost through down-sampling. This idea may be extended to create more than two levels in the HPGNN. \

\subsection{Hierarchical Point Graph}\label{section3.2}
Within the HPGNN, a graph $G = ( P, E )$ would be constructed with $N$ points, where $P = \{p_1, p_2, ..., p_N\}$ are the set of keypoints sampled at a particular level of coarseness.
Considering two adjacent levels of coarseness, we define the low-level graph $G_L = (P_L,E_L)$, and high-level graph $G_H= (P_H,E_H)$. $P_L$ is formed from the smaller voxel size to learn fine details related to a locality. $P_H$ contains keypoints from larger voxels, to learn more global and structural features from the network. Each point is characterised by $p_u = (s_u, x_u)$, with a location vector $x_u$ and a feature vector $s_u$. Hyperparameters $d_L$ and $d_H$ are radii defining the edges between points $p_u, p_v$ that form the set of edges $E_i$ for each level $i \in (L,H) $.
\begin{equation}
E_i = \{(p_u,p_v) \ | \   \Delta_x < d_{i}\} \hspace{2mm} , \hspace{5mm} \Delta_{x} = ||x_u-x_v||_{2}. \
\label{eq:1}
\end{equation}
The two graphs $G_L$ and $G_H$ are connected based on the voxels that define their points. Each point $q$ in $P_L$ will connect with a point $p$ in $P_H$, if the voxel of $p$  encapsulates $q$. After defining two graph levels and the spatial connection between them, we construct a formalism to represent how learning is possible through these connections, with the “neural message passing” approach by Hamilton \etal \cite{GRLBook,HamiltonGraphSage}. 
A node of the network has the aim of progressively refining its feature. In each iteration $t$, it uses its current feature vector and relative coordinates to generate a message/edge feature $e$ through a learnable message generating function $f(.)$.
\begin{equation}
e^t_{uv} = f(s_u, s_v , \Delta_{x}). \hspace{0mm} 
\label{eq:2}
\end{equation}
The destination vector aggregates the messages that are received from its neighbours using an aggregator $\mathrm{Agg}(.)$ and updates its feature vector $s$ through a learnable function $g(.)$.
A separate aggregator converts features between $G_L$ and $G_H$ (down/up sampling).

\subsection{Learning in the Hierarchical Graph}\label{section3.3}
Through design decisions, we strive for better GNN learning through an architecture that supports global and local message passing.
Furthermore, an appropriate choice of aggregation functions, and strategies to handle neighbourhoods of varying density and similarity are considered. The following subsections refer to Figure \ref{fig:hpgnn} and describe the learning process within two adjacent levels, which could be generalized for $n$ layers.\\ 
\noindent\textbf{Lower Graph:} The GNN of $G_L$ has higher density by design, and most neighbours of a point correspond to the same object in the dataset. Therefore, most edges of $G_L$ connect nodes of the same label, giving a high homophily as described by Zhu \etal \cite{ZhuHomophily}. In $G_H$, two neighbour nodes usually correspond to different labels, and have less homophily. We use the $\mathrm{Max}$ function as the aggregator $\mathrm{Agg}(.)$ of features from neighbouring nodes. This has the advantage of not being prone to an over-smoothing which is common in Graph Convolution Networks with mean aggregation \cite{ChenOversmooth}. The use of $\mathrm{Max}$ function should be complemented by a preprocessing step of removing outlier points, to reduce the possibility of outlier features being aggregated into a node feature.
After a feature is aggregated, the conventional approach is to use a multi-layer perceptron (MLP) to learn the function $g(.)$ and finally add $s^\mathrm{t}$ in each iteration $t$.
\begin{equation}
s^{t+1}_{u} = g (\mathrm{Agg} \{ e^t_{uv} | (u,v) \in E_i \} ) +  s^\mathrm{t}_\mathrm{u}
\label{eq:4}
\end{equation}
for each level $i \in (L,H) $. This creates a skip connection to ease learning and gradient flow as the networks get deeper. Specifically for GNN, it encourages a point feature not to deviate too far from its previous feature.\\
After $T_1$ iterations of message passing, $G_L$ has refined its features locally. Then points in $G_H$ aggregate $G_L$ features to initialise $P_H$ node features through a "downsampling" layer. Learning in $G_H$ continues for $T_2$ iterations, after which $P_H$ points return the features to the corresponding $P_L$ points through an "upsampling" layer.\\
\noindent\textbf{Downsample:}
Consider a point $p = (s_p, x_p) \in P_H$ and the corresponding set of points $\{q_1, q_2, .., q_i,.., q_k\} \in P_L$ through which edges between $G_H$ and $G_L$ are formed. Each $q_i = (s_{q_i}, x_{q_i})$ has a learnt feature $s_{q_i}$ from the preceding $T_1$ iterations. This is concatenated with $\Delta_{x} = ||x_p-x_{q_i}||$, and a function $h(.)$ learns the edge feature $e_{p,q_i}$  between $p$ and $q_i$.\

\begin{equation}
e_{p,q_i}  =  h (s_{q_i}, \Delta_x)
\label{eq:5}
\end{equation}

The downsampled feature to $G_H$ is the aggregate of such edge features between the point $p$ and $q_i, \ i \in [1,k]$. For aggregation, we use attention weights for each edge feature following an Attentive Feature Merging (AFM) scheme \cite{VelickovicGAT}. $\alpha_i$ is the attention weight for $e_{p,q_i}$. We give more representational power for the aggregation stage in the downsampling layer, since a feature representation of a larger volume should reflect the less homophilic nature of $G_H$. The aggregated edge feature is then transformed through the MLP for the downsampling function $D(.)$. $s^0_p$ is the initial feature of the 
point $p \in P_H$.\

\begin{equation}
s^0_p = D \ (\Sigma \ ( \ \alpha_i  e_{p,q_i} \ ) \ )
\label{eq:6}
\end{equation}

\input{images/image4}

\noindent\textbf{Higher Graph:} An almost exact sequence of message passing and aggregating, initialised by features in the method shown by Eq.\ref{eq:6} is used. When compared to $G_L$, the difference is the number of iterations $T_2$ that we use for the higher level. If the HPGNN has more than two levels, as the level of $G_H$ gets higher, two neighbour nodes increasingly correspond to different labels, and have less homophily.\\
\noindent\textbf{Upsample and Re-iteration:} When $G_H$ has completed iterating and learning over a wide neighbourhood, the learnt global message of each point $p = \{s_p, x_p\} \in P_H$ is passed down as an input to the upsampling function $U(.)$.\
The output feature is concatenated with the $G_L$ feature $s_{q_i}$ at the start of the downsample layer, to form a skip connection that skips the 3-layer block of a downsample layer, a GNN, and an upsample layer. Having both higher and lower level information at the starting feature vector, the lower GNN re-iterates $T_3$ steps to refine its features. We use a similar downsampling to initialise $G_L$ features from the initialised point features from the original LiDAR cloud. A similar upsampling is used to map refined $G_L$ features back to the original points. Finally, a classifier MLP predicts classes for each point.\\
\noindent\textbf{Loss Function:}
\input{tables/result1}
In the classifier MLP, we predict a multi-class probability distribution for each point.
Since large point cloud data sets have class distributions that are severely imbalanced, we use a weighted cross entropy loss $L_\mathrm{wce}$ following \cite{AlonsoWCE}. For each class $c$, a weight $w$ is defined inversely proportional to the relative frequency of the class. When $w_{c}$ and $\hat{y}_i$ are the weight and predicted probability of the labelled class, at a point $i$ in a point cloud with $N$ points, we have
\begin{equation}
L_\mathrm{wce} = -\frac{1}{N}\sum_{i=1}^{N} w_{c}\ln(\hat{y}_i).
\label{eq:8}
\end{equation}
To further improve the mean Intersection-over-Union (mIoU) of the model using the Lovasz extension of the Jaccard loss, we follow \cite{BermanLovasSoftmax} by using the Lovasz-softmax loss $L_\mathrm{ls}$. Adding an $L_{2}$ regularisation term $L_\mathrm{reg}$, the total loss for semantic segmentation is
\begin{equation}
L_\mathrm{total} = \alpha L_\mathrm{wce} \ + \beta L_\mathrm{ls} \ + \gamma L_\mathrm{reg} \ .
\label{eq:7}
\end{equation}
where $\alpha,\beta,\gamma$ are weights for each loss component.

%% file: images/image2.tex
\begin{figure}[t]
\centering
   \includegraphics[width=1\linewidth]{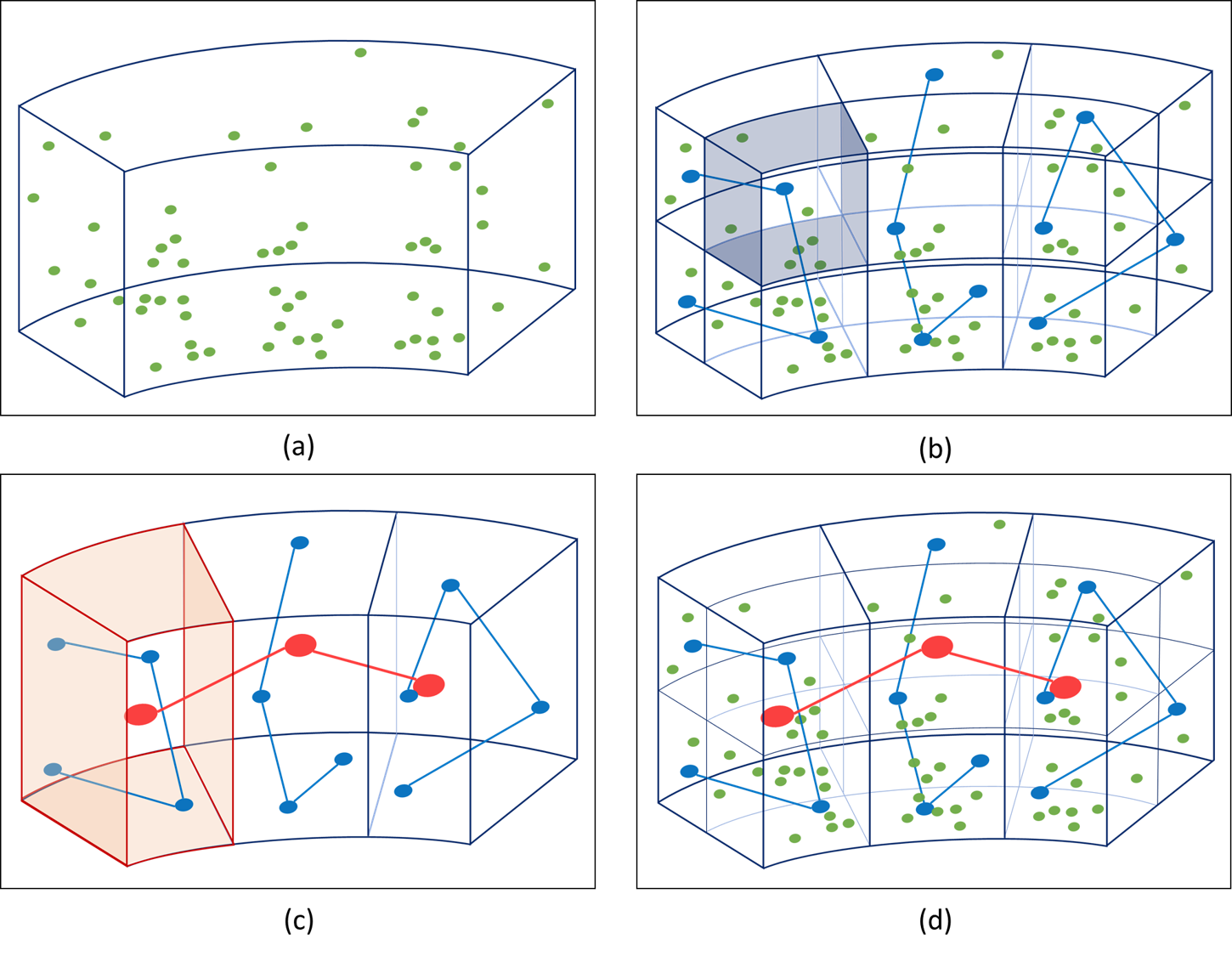}
   \caption{Graph construction by cylindrical partitioning. The dense point cloud (a) is voxelized at two levels to form a (blue) dense graph (b) and a (red) coarse graph (c). The two resulting GNNs are linked to pass information (d).}
\label{fig:vox}
\end{figure}

%% file: images/image3.tex
\begin{figure*}[t]
\centering
   \includegraphics[width=0.9\linewidth]{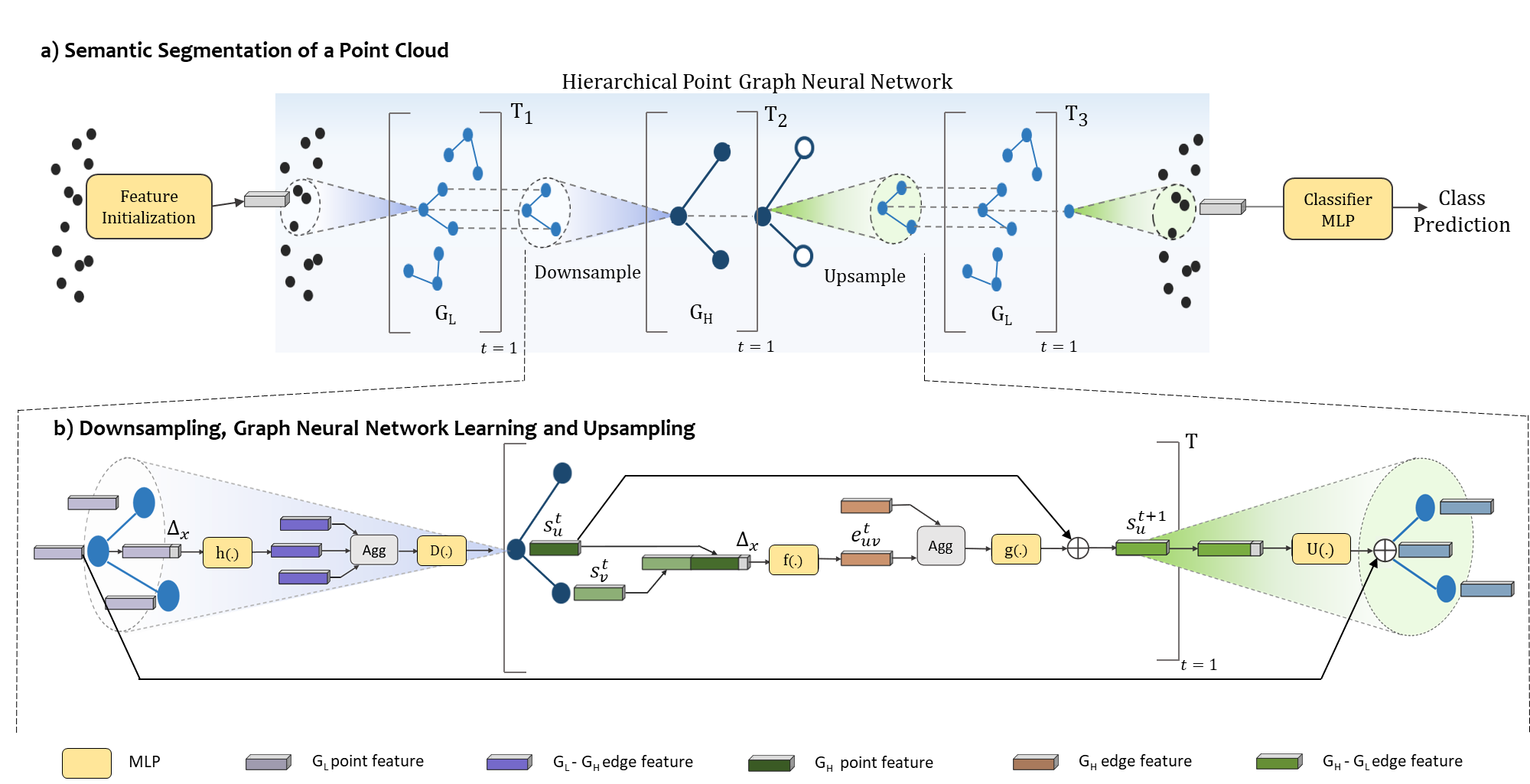} 
   \caption{a) Overview of the HPGNN architecture with feature learning in the low-level graph ($G_L$) and the high-level graph ($G_H$). b) A downsampling layer (blue) that maps local features extracted from points in $G_L$ to $G_H$, and an upsampling layer (green) that maps the learnt features from $G_H$ back to $G_L$ for classification. A skip connection adds the input feature for the three layer block (Downsampling, $G_H$, and Upsampling) to its output.}
\label{fig:hpgnn}
\end{figure*}


%% file: images/image4.tex
\begin{figure*}[t]
\centering
    \includegraphics[width=0.7\linewidth]{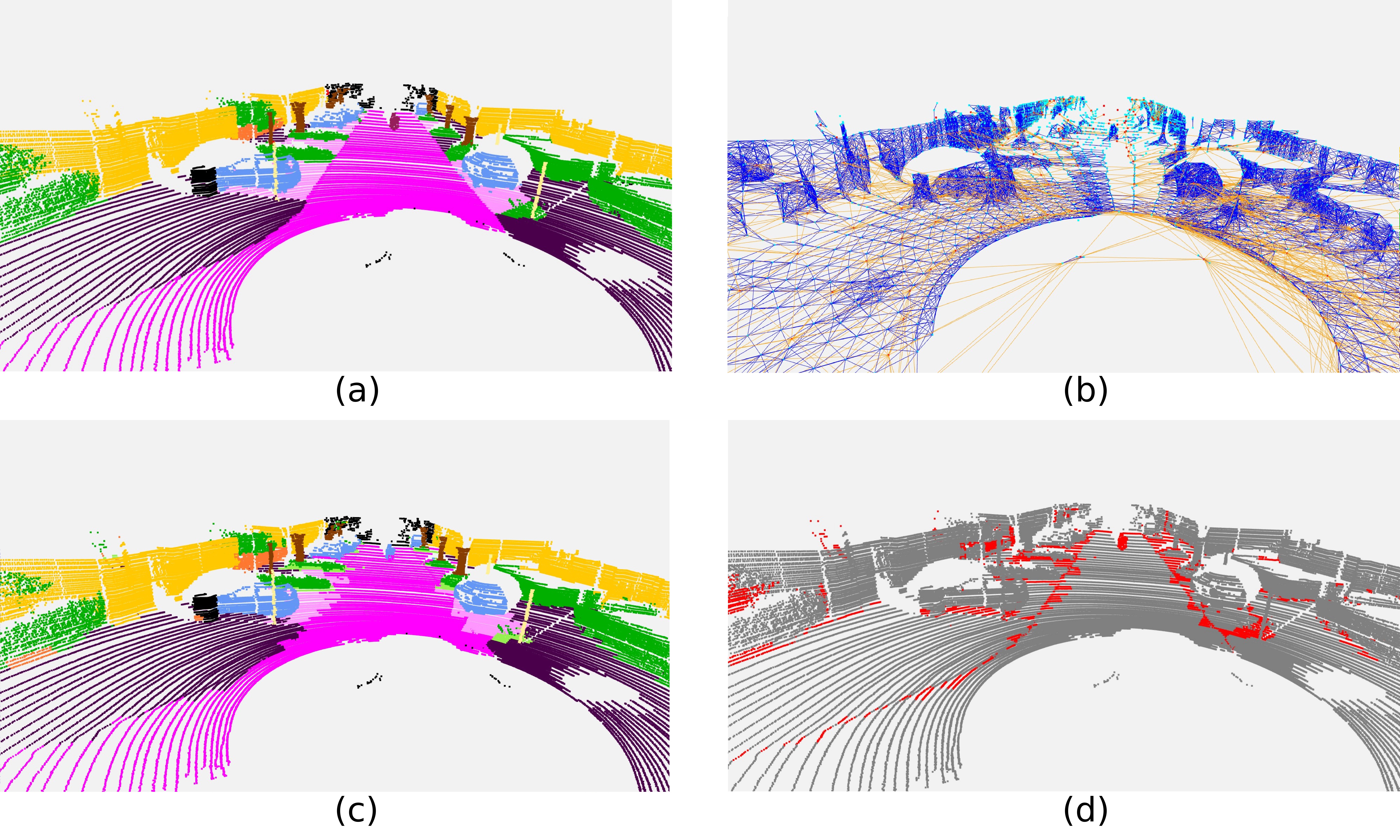} 
    \caption{(a) Ground truth labelled point cloud. (b) Construction of a HPGNN at two levels. The lower level and higher level are blue and orange respectively. (c) Segmentation output of the 2-level HPGNN. (d) Difference between the labels and predictions.\\
    \centering
    \crule[rgb_color_unlabeled]{2mm}{2mm} unlabelled
    \crule[rgb_color_car]{2mm}{2mm} car
    \crule[rgb_color_motorcyclist]{2mm}{2mm} motor-cyclist
    \crule[rgb_color_road]{2mm}{2mm} road
    \crule[rgb_color_traffic-sign]{2mm}{2mm} traffic-sign
    \crule[rgb_color_sidewalk]{2mm}{2mm} sidewalk
    \crule[rgb_color_building]{2mm}{2mm} building
    \crule[rgb_color_parking]{2mm}{2mm} parking
    \crule[rgb_color_fence]{2mm}{2mm} fence
    \crule[rgb_color_pole]{2mm}{2mm} pole
    \crule[rgb_color_vegetation]{2mm}{2mm} vegetation
    \crule[rgb_color_terrain]{2mm}{2mm} terrain
    \crule[rgb_color_trunk]{2mm}{2mm} trunk}
\label{fig:pts}
\end{figure*}

%% file: tables/result1.tex
\begin{table*}[!t]
\caption{Mean Intersection over Union (mIoU) scores compared with single scan semantic segmentation among point-based and GNN models. Our proposed model significantly outperforms existing graph based method (SP Graph) by a margin of 36.7\% mIoU, while performing competitively with the Point based State of the art.}
\label{tab:result1}
\centering
\resizebox{\textwidth}{!}{%
\begin{tabular}{|c|c|ccccccccccccccccccc|} 
\hline
Method & mIoU & \begin{sideways}Car\end{sideways} & \begin{sideways}Bicycle\end{sideways} & \begin{sideways}Motorcycle\end{sideways} & \begin{sideways}Truck\end{sideways} & \begin{sideways}Other-vehicles\end{sideways} & \begin{sideways}Person\end{sideways} & \begin{sideways}Bicyclist\end{sideways} & \begin{sideways}Motorcyclist\end{sideways} & \begin{sideways}Road\end{sideways} & \begin{sideways}Parking\end{sideways} & \begin{sideways}Sidewalk\end{sideways} & \begin{sideways}Other-ground\end{sideways} & \begin{sideways}Building\end{sideways} & \begin{sideways}Fence\end{sideways} & \begin{sideways}Vegetation\end{sideways} & \begin{sideways}Trunk\end{sideways} & \begin{sideways}Terrain\end{sideways} & \begin{sideways}Pole\end{sideways} & \begin{sideways}Traffic-sign\end{sideways} \\ 
\hline
PointNet \cite{QiPointNet}& 14.6 & 46.3 & 1.3 & 0.3 & 0.1 & 0.8 & 0.2 & 0.2 & 0.0 & 61.6 & 15.8 & 35.7 & 1.4 & 41.4 & 12.9 & 31.0 & 4.6 & 17.6 & 2.4 & 3.7 \\
PointNet++ \cite{QiPointNet+}  & 20.1 & 53.7 & 1.9 & 0.2 & 0.9 & 0.2 & 0.9 & 1.0 & 0.0 & 72.0 & 18.7 & 41.8 & 5.6 & 62.3 & 16.9 & 46.5 & 13.8 & 30.0 & 6.0 & 8.9 \\
SPLATNet \cite{Splatnet_Su_2018_CVPR} & 22.8 & 66.6 & 0.0 & 0.0 & 0.0 & 0.0 & 0.0 & 0.0 & 0.0 & 70.4 & 0.8 & 41.5 & 0.0 & 68.7 & 27.8 & 72.3 & 35.9 & 35.8 & 13.8 & 0.0 \\
TangentConv \cite{TangentTatarchenko_2018_CVPR} & 35.9 & 86.8 & 1.3 & 12.7 & 11.6 & 10.2 & 17.1 & 20.2 & 0.5 & 82.9 & 15.2 & 61.7 & 9.0 & 82.8 & 44.2 & 75.5 & 42.5 & 55.5 & 30.2 & 22.2 \\
$P^{2}$Net \cite{Li2020ProjectedpointbasedSA} & 39.8 & 85.6 & 20.4 & 14.4 & 14.4 & 11.5 & 16.9 & 24.9 & 5.9 & 87.8 & 47.5 & 67.3 & 7.3 & 77.9 & 43.4 & 72.5 & 36.5 & 60.8 & 22.8 & 38.2 \\
RandLA-Net \cite{huRandlanet} & 53.9 & \textbf{94.2} & 26.0 & 25.8 & \textbf{40.1} & \textbf{38.9} & \textbf{49.2} & 48.2 & 7.2 & \textbf{90.7} & \textbf{60.3} & \textbf{73.7} & 20.4 & \textbf{86.9} & \textbf{56.3} & \textbf{81.4} & 61.3 & \textbf{66.8} & 49.2 & 47.7 \\
\hline
\multicolumn{1}{c}{} & \multicolumn{1}{c}{} &  &  &  &  &  &  &  &  & \multicolumn{1}{l}{} & \multicolumn{1}{l}{} & \multicolumn{1}{l}{} & \multicolumn{1}{l}{} & \multicolumn{1}{l}{} & \multicolumn{1}{l}{} & \multicolumn{1}{l}{} & \multicolumn{1}{l}{} & \multicolumn{1}{l}{} & \multicolumn{1}{l}{} & \multicolumn{1}{l}{} \\ 
\hline
SPGraph \cite{LandrieuSuperPoint} & 17.4 & 49.3 & 0.2 & 0.2 & 0.1 & 0.8 & 0.3 & 2.7 & 0.1 & 45.0 & 0.6 & 34.8 & 0.6 & 64.3 & 20.8 & 48.9 & 27.2 & 24.6 & 15.9 & 0.8 \\
\textbf{HPGNN - Ours} & \textbf{54.1} & 92.7 & \textbf{33.2} & \textbf{32.2} & 26.4 & 29.4 & 43.3 & \textbf{60.9} & \textbf{19.4} & 86.3 & 50.7 & 67.5 & \textbf{27.6} & 86.8 & 55.5 & 80.7 & \textbf{61.4} & 64.1 & \textbf{50.6} & \textbf{59.1} \\
\hline
\end{tabular}}
\end{table*}

%% file: HPGNN/3_Experiments_n_Discussions.tex
\section{Experiments and Discussion}\label{section4}

\input{tables/result2}
\input{tables/result4}

To evaluate if the expected improvements in point cloud learning capability is achieved, we choose semantic segmentation as the point cloud processing task and use a 2-level HPGNN with $(T_1,T_2,T_3) = (1,2,1)$ as the hierarchical model. Following our motivation on outdoor point cloud processing, we evaluate on large scale autonomous driving datasets---SemanticKITTI \cite{behleySemanticKITTI} and nuScenes-Lidarseg \cite{nuScenes2020}. Evaluations follow mIoU recommended by \cite{behleySemanticKITTI} as the evaluation metric.\\

Figure \ref{fig:pts} depicts the process of evaluation on large scale point clouds. (a) The point-wise labelled LiDAR scan. (b) An HPGNN, with  $G_L$ and  $G_H$. $G_L$ forms most of its edges within object classes, and some edges at boundaries. $G_H$ forms edges primarily between classes, and facilitates learning over a larger distance in few iterations. The result of segmentation is in (c), with all the classes being identified, but with some edges between classes showing discretization errors.\

To compare our results, we identify two broad categories of implementations that have been used for point cloud learning tasks. The first include point-level networks, and the graph-based methods that have used information at that scale. The main motivation in these models is to use an efficient information aggregating mechanism, and not specifically focused only on semantic segmentation. The second category, are the "second generation models" that extend the point network framework to specifically focus on semantic segmentation.

\subsection{Results on SemanticKITTI}\label{section4.3}

Comparisons of mIoU scores on the SemanticKITTI benchmark with point-based and second generation models are shown in Table \ref{tab:result1} and Table \ref{tab:result2}.

\noindent\textbf{Point-based Models:} Earlier methods (PointNet\cite{QiPointNet}, PointNet++\cite{QiPointNet+}) use MLPs without a predefined underlying structure. More recent models use ideas of GNNs where an underlying structure is generated among points, and some form of information aggregation/pooling processes information from a fixed locality in each iteration (SplatNet\cite{Splatnet_Su_2018_CVPR}, TangentConv\cite{TangentTatarchenko_2018_CVPR}). These methods perform well for indoor data sets but fail to retain local information, while increasing the receptive field for large outdoor clouds. This is evident from the class-wise mIoU scores in Table \ref{tab:result1} where in pursuit of widening the receptive field by many iterations, fine information that helps to classify smaller objects are lost. In addition, they suffer from scalability issues for large point clouds as mentioned in Section \ref{section1}. RandLaNet\cite{huRandlanet} attempts to solve this issue by random sampling points, but it does not solve the issue of information loss.\

\input{tables/ablation}

SPGraph \cite{LandrieuSuperPoint} is the highest performing Graph Neural Network model that has previously attempted semantic segmentation. Although previously mentioned point based models use ideas similar to GNNs, they focus on encoding and attentive pooling modules rather than neighborhood feature aggregation mechanisms.
From the experiments it can be seen that by using the hierarchical structure for learning, we can retain local information to segment and differentiate between relatively smaller objects (e.g., motor-cyclist, bicyclist, and pole) as well as larger objects (e.g., car, terrain, and trunk). This shows our design decisions of preserving fine features while graph coarsening has been effective. Therefore, using MLPs in hierarchical levels greatly improves the baseline performance for the Graph-Neural Network implementation of the point-based learning scheme. To this end, we believe we are successful in improving the baseline of point-based GNNs that can be used for large scale point cloud processing.\\

\vspace{-2.5mm}
\noindent\textbf{Second Generation Semantic Segmentation Models:} These models in Table \ref{tab:result2} focus on the particular task of semantic segmentation and use modules that are specially tuned to improve the performance metrics. They stand upon earlier mentioned point networks by adding range based transformations (SqueezeSeg\cite{HuSqueezeandexcitation}, SalsaNext\cite{cortinhal2020salsanext}), adaptive 3D space partitions (AF2S3Net\cite{cheng2021af2s3net}, Cylinder3D\cite{ZhuCylindricalAsym}, Polarnet\cite{PolarnetZhang2020}), and 
multi view fusion, incorporating information such as bird-eye-view and range-images (MINet \cite{MINet2022}, AMVNet\cite{Liong2020AMVNetAM}, SPVNAS\cite{tang2020spvnas}, RPVNet\cite{xu2021rpvnet}). For semantic segmentation, fusion-based approaches currently achieve state-of-the-art results. RPVNet, for example, uses a three-branch fusion network. Instead of a branch with the point graph which uses a simplified PointNet \cite{QiPointNet}, an HPGNN could be used to learn additional features at various degrees of coarseness. Since we observe that HPGNN outperforms PointNet in Table \ref{tab:result1}, we can expect an improvement in second generation models by employing an HPGNN.

It is noteworthy that our model has not used any of the above enhancements specifically tailored for improving semantic segmentation. We have used hierarchical learning to improve the underlying effectiveness of message passing inside a GNN.

\subsection{Results on nuScenes-Lidarseg}\label{section4.4}
 In 2021, nuScenes-Lidarseg was released \cite{fong2021panopticnuscenes}, with point-wise annotations making it suitable for semantic segmentation. Therefore, to our knowledge, point-based models mentioned in Section \ref{section4.3} have not been evaluated in this test bench publicly. The only available comparisons are with the second generation models. They use point-based models in their feature representation step. (e.g., second generation model PolarNet \cite{PolarnetZhang2020} trains a PointNet\cite{QiPointNet} upon which ``ring convolutions" are operated.) Therefore, if our HPGNN, without any additional module, is performing competitively with such second generation models, as seen from Table \ref{tab:result4}, we infer it is an improvement of the underlying point-based model. 
This validates the generalizability of HPGNN for large scale outdoor LiDAR point clouds.

\subsection{Ablation Studies}\label{section4.5}
 Ablation studies are done using SemanticKITTI with the standard practice \cite{behleySemanticKITTI,Su3DCNN} of reserving sequence 08 of the dataset for validations.

\noindent\textbf{Hierarchical Levels:}
To validate the need of multiple levels, two models with a single level GNN each are compared with the baseline, which is a 2-level HPGNN. The first model only runs GNN on $G_L$:  $(T_1, T_2, T_3)=(1,0,1)$ and second model only runs GNN on $G_H$ graph: $(T_1, T_2, T_3)=(0,2,0)$. From Table \ref{tab:ablation}, it can be seen that a 2-level HPGNN on both $G_L$ and $G_H$ results in up-to $10.5$ mIoU improvement over single level HPGNNs. Clearly, the inability of learning at different scales hinders the performance of the model.

\noindent \textbf{GNN Iterations:} The impact of giving prominence to each level of a 2-level HPGNN through the number of allowed iterations to pass messages at each scale is examined. We analyse the contribution of increasing the receptive field of a node in $G_H$ using three models with varying $T_2$; $(T_1, T_2, T_3) \in \{(1,1,1), (1,2,1), (1,3,1)\}$, then the same effect in $G_L$ using models with $T_2$ constant. Table \ref{tab:ablation} shows that $T_2=2$ has slight improvement over $T_2=1$, and that $T_2=3$ reverts back to a lower mIoU. This might be due to the receptive field being unnecessarily large compared to the dimensions of the instances in this dataset.\

Increasing $T_1, T_3$ shows improvements of $4.1$ mIoU. This suggests that refining and learning features at $G_L$ significantly improves performance with moderate reception at $G_H$. 

\subsection{Limitations of HPGNN}\label{section4.6}
HPGNN is used as a framework for learning features from a point-based representation of a point cloud. For computational feasibility, we use voxelization. As presented by \cite{Chen2018FastRO}, spatially uniform re-sampling has discretization errors, and this is evident from Figure \ref{fig:pts} where the errors of segmentation are mostly at the boundaries.

As opposed to graph-resampling methods including Octrees \cite{Octree2005}, HPGNN requires dense graphs to achieve high resolution, which leads to spatial inefficiency.
This is when redundant points of a dense point cloud are stored and processed that do not contribute for evolving new features. This is a limitation in HPGNN in the graph construction step.
Although our idea of connecting local points with similar semantics for efficient representation is explored in a different approach by Hypergraphs\cite{Zhang2021HypergraphSA}, it is not entirely dependent on point distance. 

%% file: tables/result2.tex
\begin{table*}[!t]
\caption{mIoU scores compared with single scan semantic segmentation on SemanticKITTI test set. Our model performs competitively with second generation models employing point networks or graph networks as their underlying architecture.}
\label{tab:result2}
\centering
\resizebox{\textwidth}{!}{%
\begin{tabular}{|c|c|ccccccccccccccccccc|} 
\hline
Method & mIoU & \begin{sideways}Car\end{sideways} & \begin{sideways}Bicycle\end{sideways} & \begin{sideways}Motorcycle\end{sideways} & \begin{sideways}Truck\end{sideways} & \begin{sideways}Other-vehicles\end{sideways} & \begin{sideways}Person\end{sideways} & \begin{sideways}Bicyclist\end{sideways} & \begin{sideways}Motorcyclist\end{sideways} & \begin{sideways}Road\end{sideways} & \begin{sideways}Parking\end{sideways} & \begin{sideways}Sidewalk\end{sideways} & \begin{sideways}Other-ground\end{sideways} & \begin{sideways}Building\end{sideways} & \begin{sideways}Fence\end{sideways} & \begin{sideways}Vegetation\end{sideways} & \begin{sideways}Trunk\end{sideways} & \begin{sideways}Terrain\end{sideways} & \begin{sideways}Pole\end{sideways} & \begin{sideways}Traffic-sign\end{sideways} \\ 
\hline
MINet \cite{MINet2022} & 54.3 & 85.2 & 38.2 & 32.1 & 29.3 & 23.1 & 47.6 & 46.8 & 24.5 & 90.5 & 58.8 & 72.1 & 25.9 & 82.2 & 49.5 & 78.8 & 52.5 & 65.4 & 37.7 & 55.5 \\
PolarNet \cite{PolarnetZhang2020} & 54.3 & 93.8 & 40.3 & 30.1 & 22.9 & 28.5 & 43.2 & 50.2 & 5.6 & 90.8 & 61.7 & 74.4 & 21.7 & 90.0 & 61.3 & 84.0 & 65.5 & 67.8 & 51.8 & 57.5 \\
SqueezeSegv3 \cite{xu2021squeezesegv3} & 55.9 & 92.5 & 38.7 & 36.5 & 29.6 & 33.0 & 45.6 & 46.2 & 20.1 & 91.7 & 63.4 & 74.8 & 26.4 & 89.0 & 59.4 & 82.0 & 58.7 & 65.4 & 49.6 & 58.9 \\

 
Cylinder3D \cite{ZhuCylindricalAsym} & 67.8 & 97.1 & 67.6 & 64.0 & \textbf{59.0} & 58.6 & 73.9 & 67.9 & 36.0 & 91.4 & 65.1 & 75.5 & 32.3 & 91.0 & 66.5 & 85.4 & 71.8 & 68.5 & 62.6 & 65.6 \\
AF2S3Net \cite{cheng2021af2s3net} & 69.7 & 94.5 & 65.4 & \textbf{86.8} & 39.2 & 41.1 & \textbf{80.7} & \textbf{80.4} & \textbf{74.3} & 91.3 & 68.8 & 72.5 & \textbf{53.5} & 87.9 & 63.2 & 70.2 & 68.5 & 53.7 & 61.5 & \textbf{71.0} \\ 
SPVNAS \cite{tang2020spvnas} & 67.0 & 97.2 & 50.6 & 50.4 & 56.6 & 58.0 & 67.4 & 67.1 & 50.3 & 90.2 & 67.6 & 75.4 & 21.8 & 91.6 & 66.9 & 86.1 & 73.4 & 71.0 & 64.3 & 67.3 \\
AMVNet \cite{Liong2020AMVNetAM} & 65.3 & 96.2 & 59.9 & 54.2 & 48.8 & 45.7 & 71.0 & 65.7 & 11.0 & 90.1 & \textbf{71.0} & 75.8 & 32.4 & 92.4 & 69.1 & 85.6 & 71.7 & 69.6 & 62.7 & 67.2 \\
RPVNet \cite{xu2021rpvnet} & \textbf{70.3} & \textbf{97.6} & \textbf{68.4} & 68.7 & 44.2 & \textbf{61.1} & 75.9 & 74.4 & 73.4 & \textbf{93.4} & 70.3 & \textbf{80.7} & 33.3 & \textbf{93.5} & \textbf{72.1} & \textbf{86.5} & \textbf{75.1} & \textbf{71.7} & \textbf{64.8} & 61.4 \\

\hline
\multicolumn{1}{c}{} & \multicolumn{1}{c}{} &  &  &  &  &  &  &  &  & \multicolumn{1}{l}{} & \multicolumn{1}{l}{} & \multicolumn{1}{l}{} & \multicolumn{1}{l}{} & \multicolumn{1}{l}{} & \multicolumn{1}{l}{} & \multicolumn{1}{l}{} & \multicolumn{1}{l}{} & \multicolumn{1}{l}{} & \multicolumn{1}{l}{} & \multicolumn{1}{l}{} \\ 
\hline
\textbf{HPGNN - Ours} & 54.1 & 92.7 & 33.2 & 32.2 & 26.4 & 29.4 & 43.3 & 60.9 & 19.4 & 86.3 & 50.7 & 67.5 & 27.6 & 86.8 & 55.5 & 80.7 & 61.4 & 64.1 & 50.6 & 59.1 \\
\hline
\end{tabular}}
\end{table*}

%% file: tables/result4.tex
\begin{table}[!t]
\vspace{-10pt}
\caption{mIoU scores compared with single scan semantic segmentation on nuScenes-Lidarseg test set.}
\label{tab:result4}
\centering

\begin{tabular}{|c|c|} 
\hline
Method & mIoU\\ 
\hline
MINet \cite{MINet2022} & 56.4\\

PolarNet \cite{PolarnetZhang2020} & 69.4\\

Cylinder3D \cite{ZhuCylindricalAsym} & 77.2\\

AMVNet \cite{Liong2020AMVNetAM} & 77.3\\

SPVNAS \cite{tang2020spvnas} & 77.4\\

AF2S3Net \cite{cheng2021af2s3net} & 78.3\\

SPVCNN++ \cite{tang2020spvnas} & 81.1\\ 

\hline
\hline
\textbf{HPGNN - Ours} & 63.8\\
\hline
\end{tabular}
\end{table}

%% file: tables/ablation.tex
\begin{table}[!t]
\vspace{-10pt}
\centering
\caption{Ablation study on the need for multiple levels and number of iterations in each level.}
\label{tab:ablation}
\begin{tabular}{|l|l|l|l|}
\hline
T1 & T2 & T3 & mIoU \\ 
\hline
1 & 0 & 1 & 43.3 \\
0 & 2 & 0 & 41.0 \\
1 & 1 & 1 & \textbf{51.5}\\
\hline
1 & 2 & 1 & 51.6\\
1 & 3 & 1 & 48.5\\
2 & 1 & 2 & 53.4\\
2 & 2 & 2 & 53.1\\
3 & 1 & 3 & \textbf{55.6}\\
\hline
\end{tabular}
\end{table}

%% file: HPGNN/4_Conclusion.tex
\section{Conclusion}\label{section5}
We introduced a novel framework for learning discriminative features of large-scale outdoor point clouds, through graph neural message passing.
The hierarchical point graph neural network (HPGNN) learns features of various levels of coarseness from point clouds. Our experiments show that it is a resource efficient method for expanding the effective receptive field of each point in a graphical structure. HPGNN was designed as a GNN based approach, and it significantly improved the existing baseline of GNN based point cloud segmentation (+36.7 mIoU on SemanticKITTI dataset). Our results suggest an improved GNN based backbone for point cloud processing, which may be extended upon by modular designs or fusion-based approaches for semantic segmentation and other tasks related to point clouds.\footnote{The code will be made available here: \href{https://git.io/JVhDc}{https://git.io/JVhDc}}.\\

%% file: HPGNN/Acknowledgement.tex
\noindent{Acknowledgement}: We thank National Research Council of Sri Lanka for providing computational resources through the grant no. 19-080.